# Automated Dynamic Image Analysis for Particle Size and Shape Classification in Three Dimensions

Sadegh Nadimi[1,2], Vasileios Angelidakis[3,2], Sadaf Maramizonouz[1], Chao Zhang[1]

[1] School of Engineering, Newcastle University, Newcastle upon Tyne, United Kingdom

[2] Ocular Systems Ltd., Newcastle upon Tyne, United Kingdom

[3] School of Natural and Built Environment, Queen's University Belfast, Belfast, United Kingdom

**ABSTRACT**

We introduce OCULAR, an innovative hardware and software solution for three-dimensional dynamic image analysis of fine particles. Current state-of-the art instruments for dynamic image analysis are largely limited to two-dimensional imaging. However, extensive literature has demonstrated that relying on a single two-dimensional projection for particle characterisation can lead to inaccuracies in many applications. Existing three-dimensional imaging technologies, such as computed tomography, laser scanning, and orthophotography, are limited to static objects. These methods are often not statistically representative and come with significant post-processing requirements, as well as the need for specialised imaging and computing resources. OCULAR addresses these challenges by providing a cost-effective solution for imaging continuous particle streams using a synchronised array of optical cameras. Particle shape characterisation is achieved through the reconstruction of their three-dimensional surfaces. This paper details the OCULAR methodology, evaluates its reproducibility, and compares its results against X-ray micro computed tomography, highlighting its potential for efficient and reliable particle analysis.

Correspondence: sadegh.nadimi-shahraki@ncl.ac.uk

## Introduction

Particle size and shape are essential characteristics of particulate materials significantly influencing their behaviours and interactions with their surroundings across diverse applications such as food industry [1, 2], cell biology [3], pharmaceutical tableting [4], transport [5], and even space exploration [6].

Conventionally, particle shape characterisation has relied on two-dimensional (2D) analysis of particle projections using microscopy techniques. While 2D shape characterisation could provide valuable insights particularly for general shape or size distribution, it may not be adequate as it inherently overlooks the particle third dimension causing unrealistic assumptions. This can result in limitations when translating findings from 2D representations to real-world applications particularly where the full 3D shape impacts material interactions, flow dynamics, or structural stability [7].

In response, three-dimensional (3D) characterisation has emerged as a more comprehensive method [8, 9]. By capturing the full morphology, 3D particle shape characterisation enables a deeper understanding of particle behaviour in various contexts. The technologies currently available for 3D imaging such as computed tomography, laser-scanning, and orthophotography offer high accuracy but come with considerable challenges. These systems require substantial investment in both equipment and data processing time and are often limited to static and small sample sizes, restricting their applicability for real-time analyses that demand rapid, large-scale assessments, such as in processing or production settings [7].

To address these limitations, the current work proposes a novel and cost-effective hardware and software solution for the 3D dynamic image acquisition and shape characterisation of particles. In this system a continuous flow of a fluid-particles suspension is recorded using a set of synchronised optical cameras. Then, particle morphology, i.e. size and shape, is characterised by reconstructing the real 3D particle geometry. The present setup delivers a more realistic representation of particle morphology compared to 2D dynamic image analysis solutions, enhancing the accuracy and applicability of shape characterisations across various fields involving fine particles.

## Methods

*Components of OCULAR*

OCULAR comprises optical imaging components, connected to and controlled by a central computing unit which is also used for synchronised image acquisition of the 2D projections, for 3D reconstruction and subsequently for particle characterisation and classification. In the current work, three orthogonally positioned optical cameras (IDS uEye) each connected with a lens (IDS-U3-3040CP-M-GL-R2) mounted on a rig are connected to an interface card of a workstation via USB 3.0 cables. The region of interest (RoI) for the imaging is illuminated by two LED lights, a rectangular LED light placed at the back of the camera rig and an annular LED light placed below a capillary tube. The tube (Hollow Square Capillaries, Part Number 8100-100, CM Scientific, UK) with a 1 mm × 1 mm square cross-section (square capillary) and a 100 mm length is placed in front of the three cameras by mounting it on the same rig. The square capillary is connected to a syringe pump (Aladdin AL-1000 SyringeONE Programmable Syringe Pump) using flexible tubing (RS PRO Silicone Stock No.273-2490, RS Components Ltd, UK) with inner and outer diameters of 1.5 mm and 2.5 mm, respectively. **Figure 1** shows an outline of the OCULAR set-up.

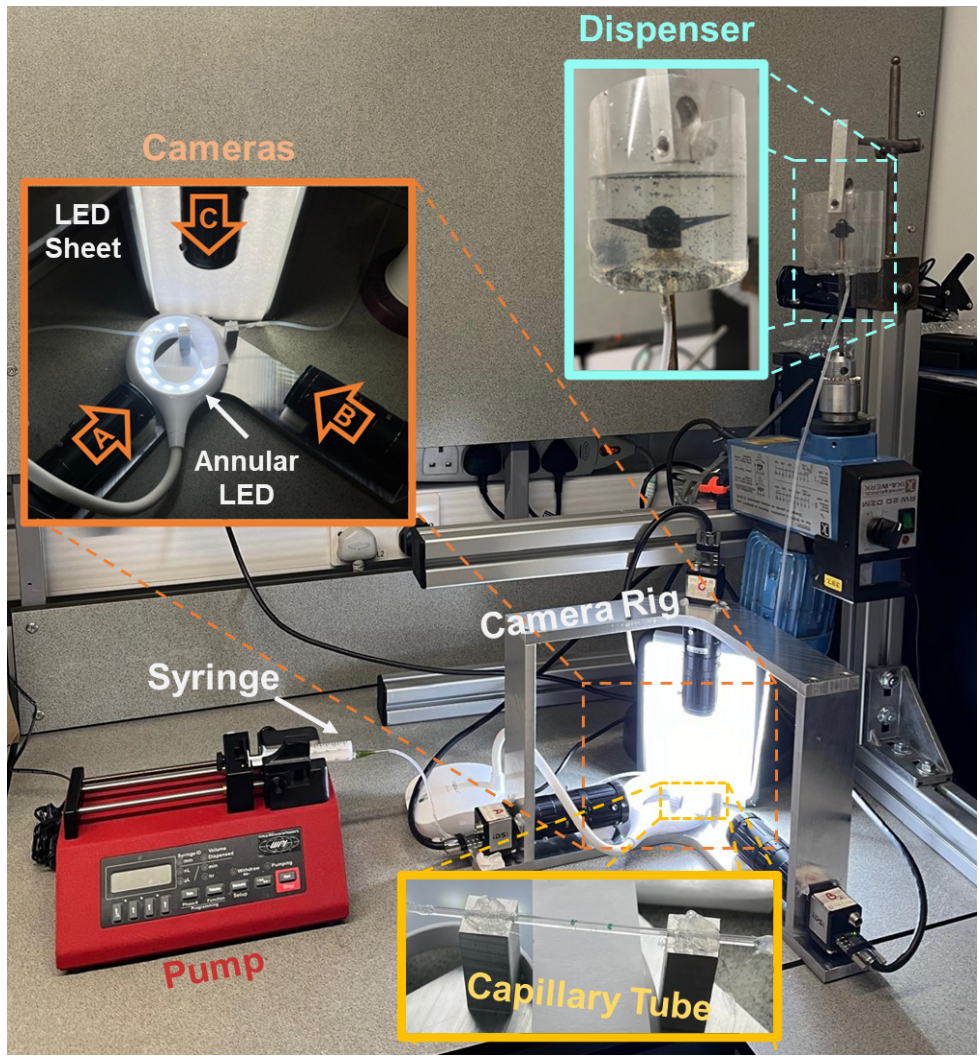

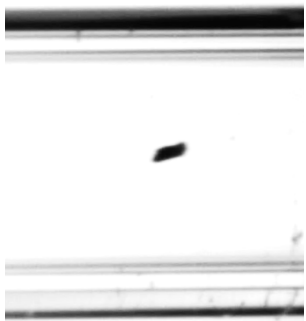
**Camera A**

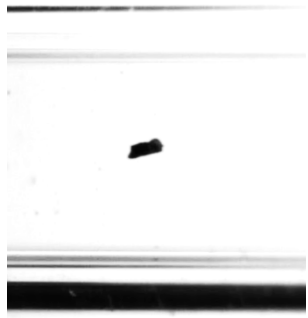
**Camera B**

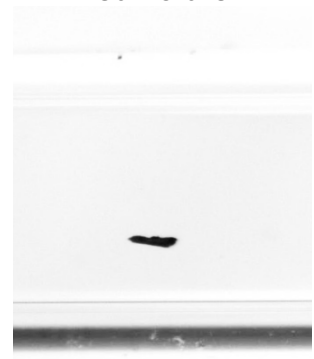
**Camera C**

**Figure 1** An outline of the OCULAR set-up, demonstrating the camera arrangement, dispensing unit, illumination, and the capillary tube.

*Image acquisition of dynamic objects*

A fluid-particle suspension is made from mixing alumina particles (L.B. Foster Rail Technologies, UK) in a solution of 0.1 g of xanthan powder (Xanthan gum from *Xanthomonas campestris,* Sigma Aldrich, UK) to 100 g deionised (DI) water. The suspension is pumped and/or withdrawn through the fluidic system using the syringe pump with a set flow rate. Two containers are utilised at each of the fluidic system: one for containing the fluid-particle sample which is constantly stirred using a mixer (RW 20 DZM, IKA Werke GmbH Janke & Kunkel, Germany) to avoid sedimentation of the alumina particles, and the other for collecting the output of the system (**Figure 1**). The fluid-particle suspension flows through the flexible tubing and then the square capillary. The three cameras capture synchronised videos of the three views of each particle as they enter and exit the RoI. Three synchronised 2D projections of each individual particle are extracted from the videos and then processed for the 3D reconstruction of particle shape.

*Dynamic calibration*

Before conducting the tests, the optical imaging system is calibrated by capturing three images of sapphire precision spheres (Sapphire Balls Product Code: AL66-SP-000105, Goodfellow, USA) with a diameter of 250 µm and manufacturing tolerance of 2.5 µm, which is suspended in the same xanthan-water solution and sealed inside a square capillary. The three 2D projections are then compared to the geometry of the sapphire precision spheres produced to calibrate the three cameras.

*Principles and significance of the 3D-reconstruction technique*

Each 2D projection is captured initially as a grayscale image. It is then binarised into a black and white image, and denoising is used to remove any isolated pixels, so that a single object is identified in the image, corresponding to the particle projection. Volume extrusion is performed in different directions, and the common volume is recorded.

Even though measuring the real convexity of the particle accurately is at risk following this reconstruction approach, the overall form in terms of flatness and elongation is still represented well, if a minimal bounding box is used to estimate the main particle dimensions, since this is not affected by convexity. By definition, a particle and the convex hull of the particle share the same minimal bounding box and thus main dimensions. This ability to capture particle form positions the system in significant advantage compared to 2D image characterisation instruments, which rely solely on a single projection of the particle, risking the failure of capturing both convexity and form.

A 2D image of a particle cannot represent whether a particle is compact, flat, elongated, or bladed (flat and elongated), as there is no information of what is happening in the other direction. As a result, a particle might look compact in a 2D image but be elongated in reality or look elongated in a 2D image but be flat in reality. 3D objects can only be imaged adequately using 3D information, as inferring their characteristics from a single perspective bears a significant amount of ambiguity on the actual particle geometry. Hollow particles, or particles with concealed concavities, which cannot be captured from perimetric images, can be imaged using laser scanning (if the particle orientation allows) or using X-ray micro computed tomography regardless of particle orientation, but both methods require the particle to be static for prolonged amounts of imaging time, and do not allow the imaging of continuous streams of particles. The apparatus presented here overcome the limitation of 2D imaging while offering efficient 3D analysis of particles in high throughput flow.

*How to describe size and shape in 3D*

In 3D analysis, particle size is typically characterised through dimensions along orthogonal axes, allowing for comprehensive representation of particle dimension. A common approach is using a bounding box which defines a rectangular prism tightly enclosing the particle, and capturing its short (S), intermediate (I), and long (L) dimensions in three perpendicular directions. The intermediate dimension has been used to represent the particle size. Additionally, other metrics like volume and surface area are utilised for 3D size characterisation [10].

Characterising particle shape in three dimensions requires indices that capture the morphology adequately, focusing on shape descriptors such as elongation, flatness, compactness, and sphericity. The Zingg shape classification system [11] is one of the foundational methods for categorising particle shape, utilising the values of particle flatness and elongation to classify particles into for shape categories of compact, flat, elongated, and bladed. In the Zingg chart, the elongation index is defined as the ratio of a particle intermediate to long dimensions (I/L), while the flatness index is defined as the ratio of the particle short to intermediate dimensions (S/I). By combining multiple shape descriptors, researchers can achieve a detailed and consistent representation of particle morphology, facilitating accurate modelling in various applications. A recently proposed system redefined elongation and flatness to overcome known limitation of the Zingg system [10].

**Results and Discussion**

The size and shape of 81 alumina particles are characterised using the OCULAR system and the SHAPE code [12]. **Figure 2** shows **(a)** the Particle Size Distribution (PSD) based on the particle intermediate dimension and **(b)** the shape distribution of the particles plotted on a Zingg chart [11] and coloured by the particle number density in sub-regions of the chart. It can be seen that the majority of particles belong to the elongated shape category. The particles surface area, volume, true sphericity [13], intercept sphericity [14], and convexity are presented in **Figure 3(a)** to **(e)**.

*Reproducibility of results*

The reproducibility of the OCULAR system is quantified by characterising the same fluid-particle suspension sample of alumina in a 0.2 g to 100 g solution of xanthan powder and DI water, which is prepared and run through the system three times. For each run of the same sample, the three cameras capture synchronised videos of the three views and individual particles are 3D reconstructed and characterised. The number of 182, 127, and 71 alumina particles are identified and characterised for the first, second, and third run of the experiment, respectively. A loss of 55 and 56 particles occurs between the first and second, and the second and third runs of the repeatability experiments, respectively. The particle size and shape distribution data obtained from the three runs are presented in **Figure 4** to show the reproducibility of the data presented by the OCULAR system, highlighting the loss of particles has minimal effect on the quality of measurement.

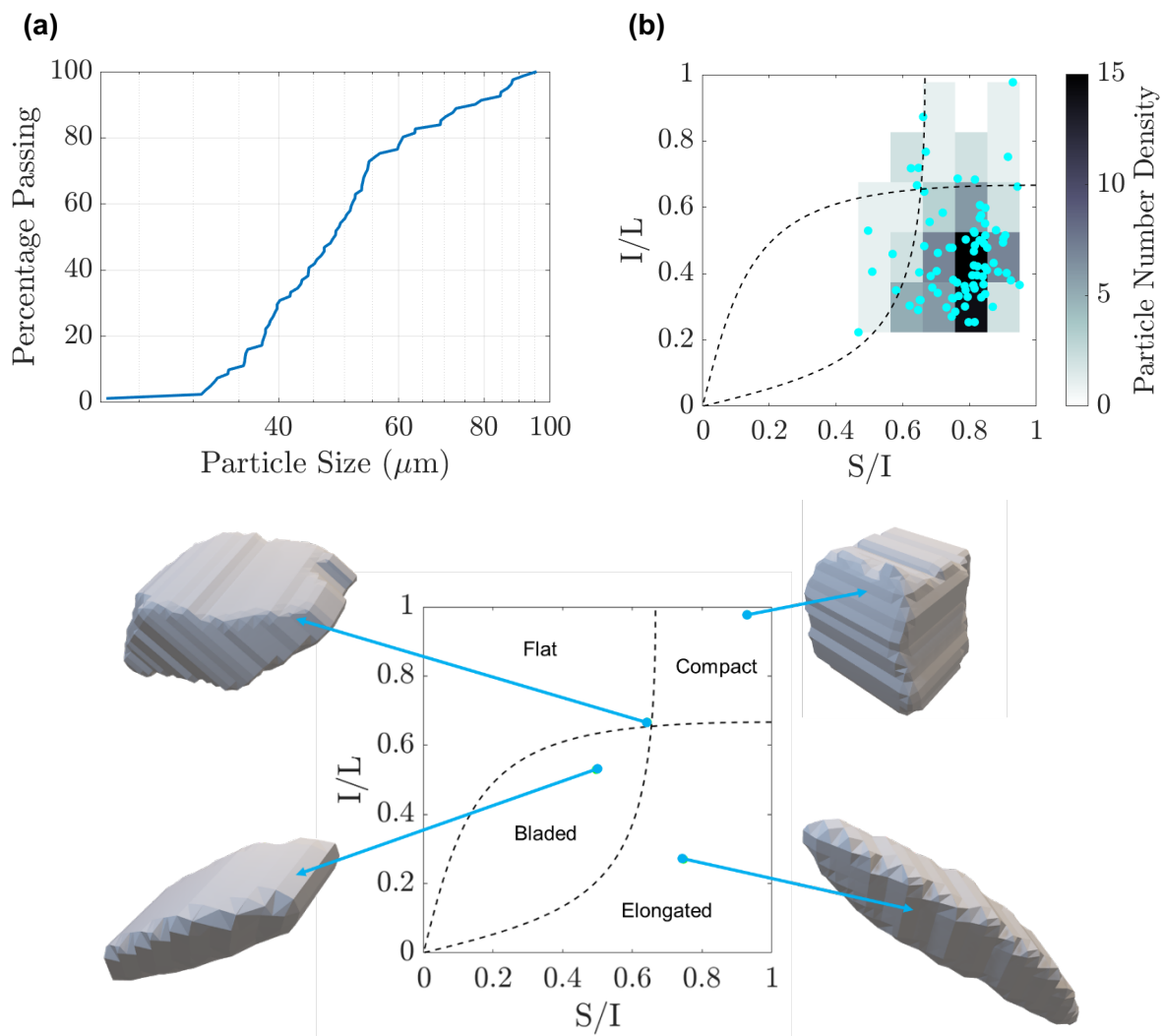

**Figure 2** The **(a)** particle size distribution of the alumina particles, and **(b)** particle shape distribution of the alumina sample presented on a Zingg [11] chart and coloured by the particle number density.

**Figure 4(a)** illustrates that the measurement of particle size distribution is highly consistent throughout the three runs of the repeatability study with a slight discrepancy for particles smaller than 40 µm. **Figure 4(b-1)** to **(b-3)** show the particle shape distributions for the three runs which confirm the consistency of the measurement, also, in terms of particle shape.

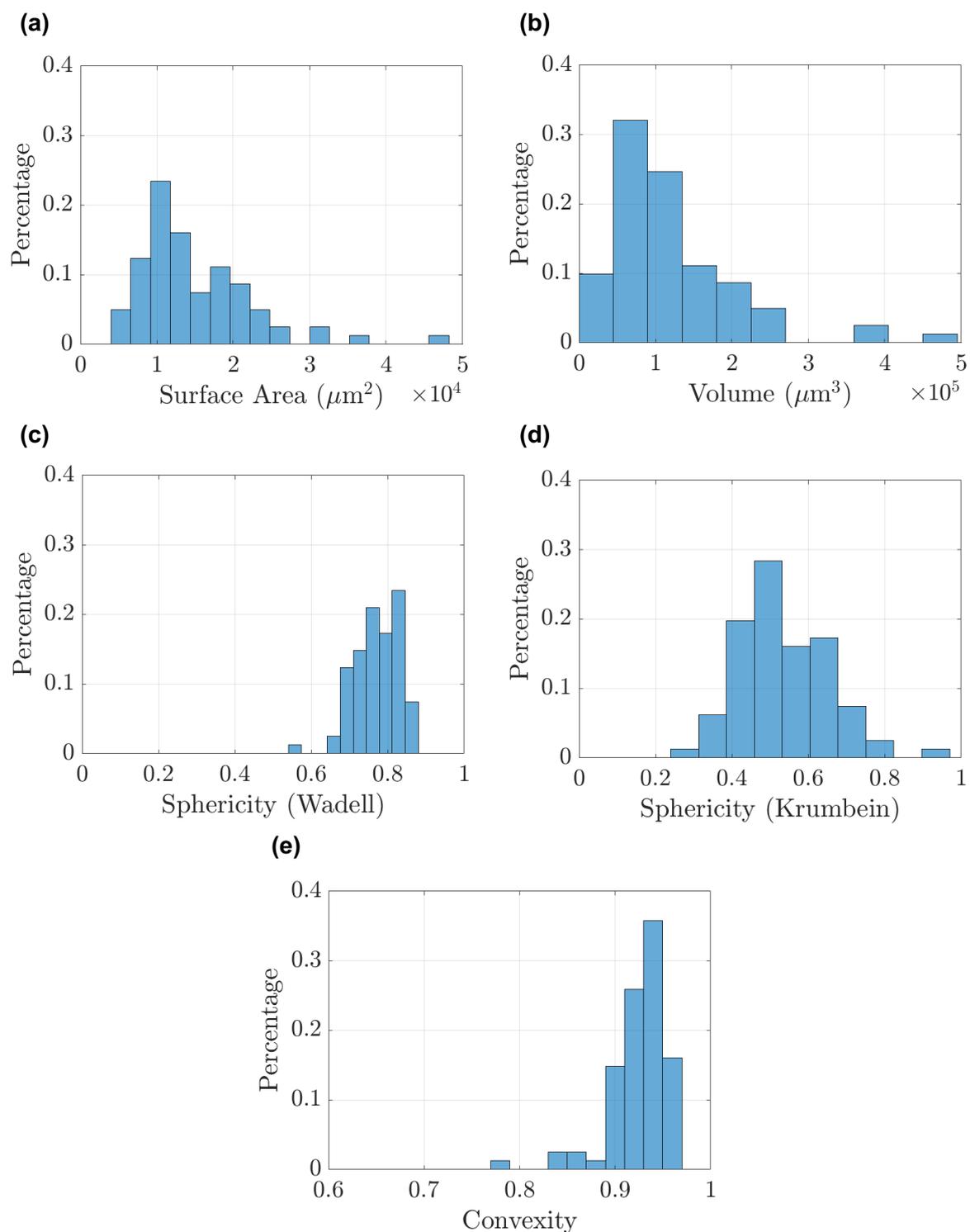

**Figure 3** The particles' **(a)** surface area, **(b)** volume, **(c)** true sphericity [13], **(d)** intercept sphericity [14], and **(e)** convexity of the alumina sample.

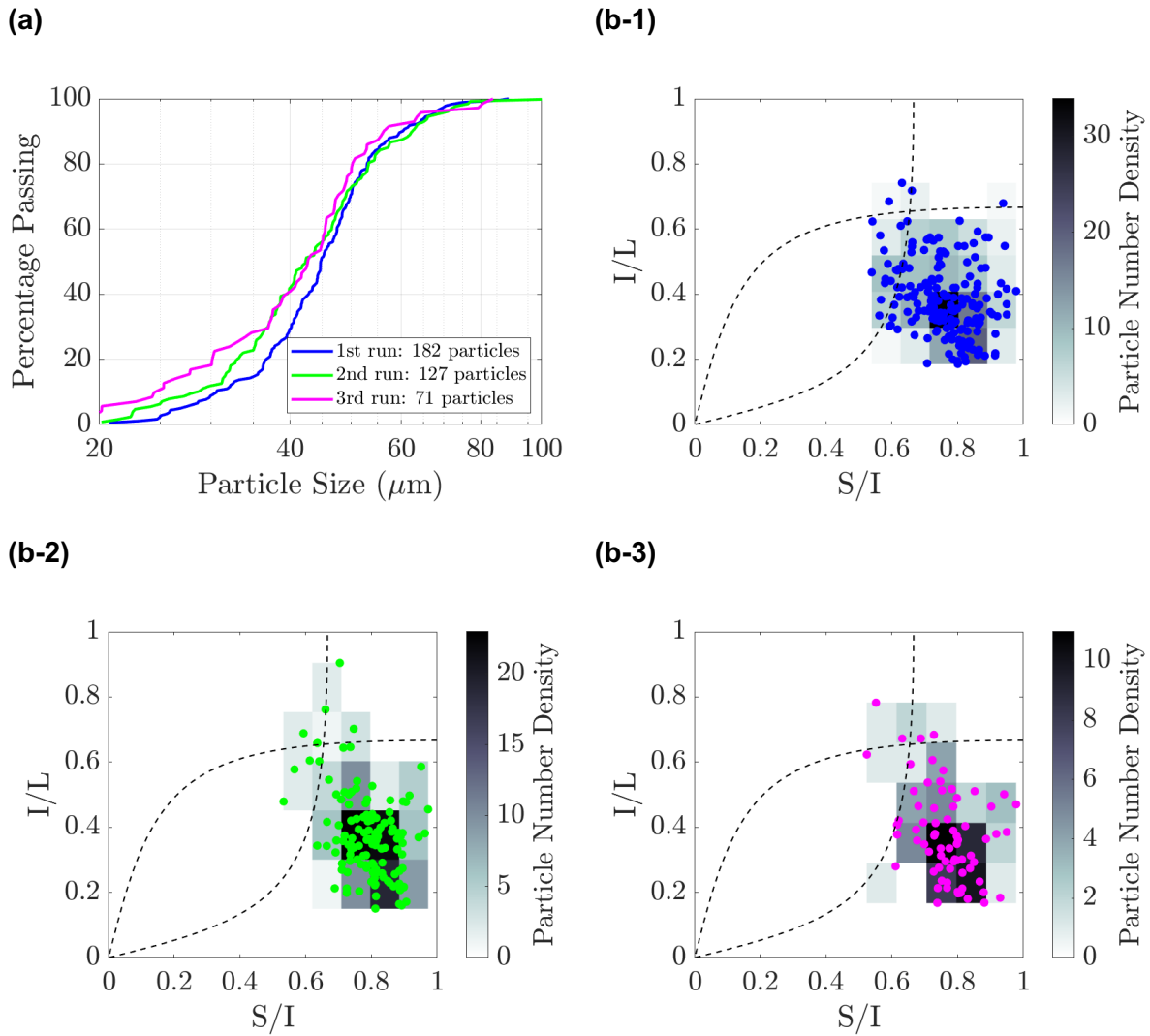

**Figure 4** The alumina particle **(a)** size distribution based on the intermediate dimension, and **(b)** shape distribution (presented on a Zingg [11] chart and coloured by the particle number density) for the **(1)** first, **(2)** second, and **(3)** third run of the same sample during the repeatability study.

*Comparison with μCT*

The size and shape distributions of 154 alumina particles, obtained through μCT and the OCULAR, are presented in **Figure 5(a)**, **(b-1)**, and **(b-2)**. Shape characterisation is performed using the SHAPE code [12]. The particle size distribution data in **Figure 5(a)** demonstrate good agreement between the μCT and OCULAR results, with the latter slightly overestimating particle sizes. **Figure 5(b-1) and (b-2)** compares the particle shape distributions from μCT and OCULAR, revealing that OCULAR provides a classification that closely matches that of μCT.

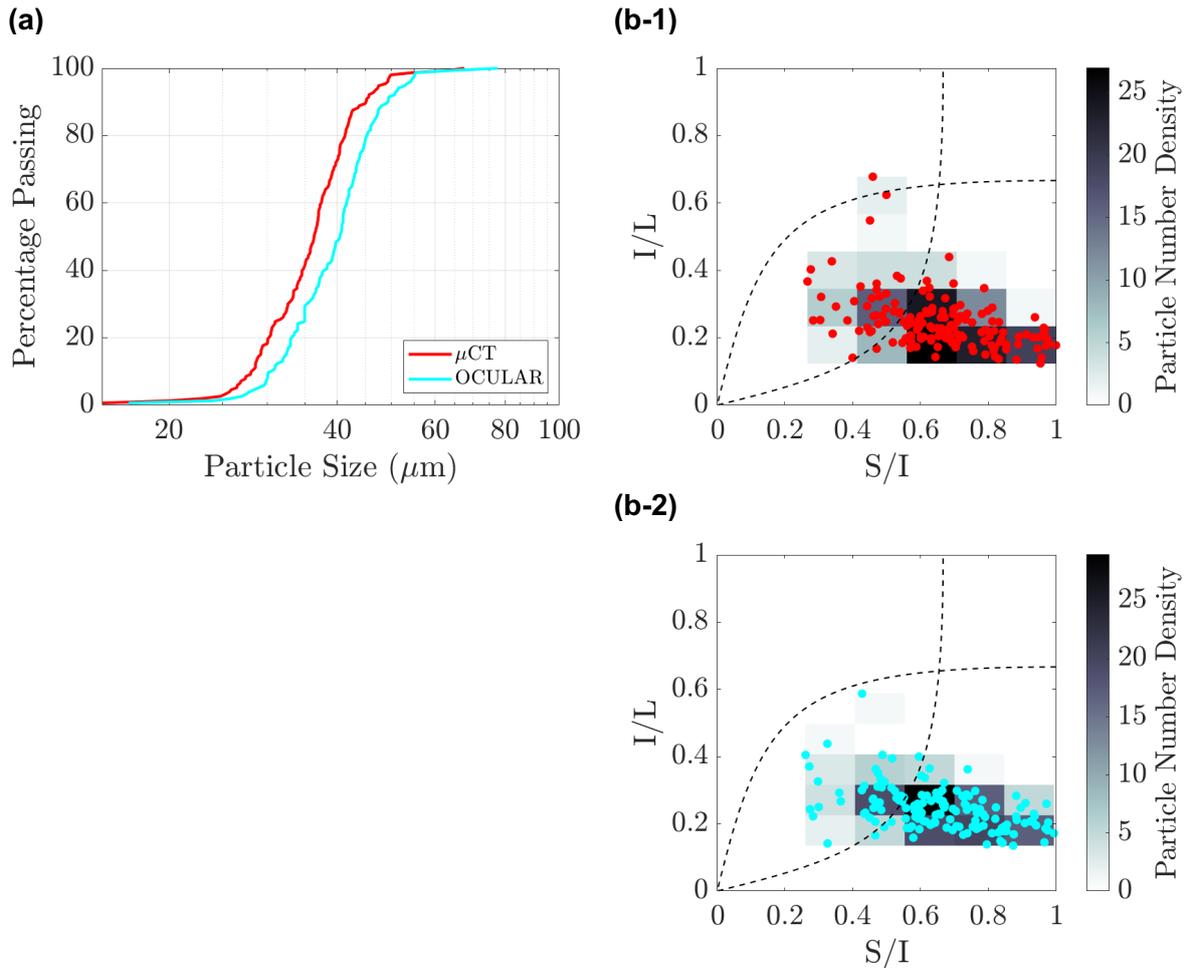

**Figure 5** The **(a)** particle size distribution of the alumina sample, and **(b)** particle shape distribution of the alumina sample presented on a Zingg [11] chart and coloured by the particle number density obtained through **(b-1)** µCT, and **(b-2)** OCULAR;

**Concluding Remarks**

This paper presented OCULAR, a hardware and software solution for three-dimensional dynamic image analysis of fine particles. By addressing the limitations of traditional two-dimensional imaging, OCULAR provides a more accurate and comprehensive method for particle shape and size characterisation. Unlike existing 3D imaging techniques, which are often limited to static samples, OCULAR enables the analysis of continuously flowing particles, making it particularly suitable for high-throughput analysis. The system ability to capture detailed 3D particle morphology through synchronized optical cameras allows for robust and real-time particle characterisation, offering significant advantages over conventional methods.

The reproducibility of the OCULAR system has been validated, showing consistent results in particle size and shape distributions across multiple experimental runs. Moreover, the comparison between OCULAR and µCT demonstrates strong agreement in both size and shape measurements, further confirming the reliability and accuracy of the system. These findings highlight the potential of OCULAR as an efficient tool for a wide range of applications involving particles.

**Acknowledgments**

This work was supported by the Engineering and Physical Sciences Research Council, grant number EP/R511584/1 entitled "OCULAR: automated acquisition and classification of particles" and Northern Accelerator, grant numbers: NACCF244 and NA-SPF11, entitled "OCULAR2.0: automated acquisition and classification of particles" and "OCULAR Smart Sample Presentation Unit", respectively.

**Conflict of Interests**

The first and second authors are founders and directors of OCULAR Systems Ltd. and have a patent for "Three-Dimensional Object Shape Acquisition Characterisation and Classification" (International Publication Number: WO 2023/180742). The remaining authors have no conflicts of interest to declare.